\documentclass[10pt,twocolumn,letterpaper]{article}

\usepackage[accsupp]{axessibility} 
\usepackage[pagenumbers]{cvpr} 
\usepackage{xcolor} 
\usepackage{algorithm}
\usepackage{algorithmic}

\newcommand{\guidemodel}{$G$}
\newcommand{\versionone}{full guide model}

\newcommand{\versiontwo}{tiny guide model}

\definecolor{darkgreen}{rgb}{0.0, 0.5, 0.0}

\newcommand{\YT}[1]{{#1}}
\newcommand{\modified}[1]{{\textcolor{gray}{}}}

\newcommand{\fref}[1]{Figure~\ref{#1}}

\newcommand{\tref}[1]{Table~\ref{#1}}

\definecolor{cvprblue}{rgb}{0.21,0.49,0.74}

\usepackage{graphicx}
\usepackage{caption}
\usepackage{float}

\usepackage{booktabs} 
\usepackage[pagebackref,breaklinks,colorlinks,citecolor=cvprblue]{hyperref}
\def\paperID{6962} 
\def\confName{CVPR}
\def\confYear{2024}

\title{Plug-and-Play Diffusion Distillation}

\author{
    Yi-Ting Hsiao\textsuperscript{1,2\thanks{Work done during internships at Adobe Inc.}},
     Siavash Khodadadeh\textsuperscript{2},
     Kevin Duarte\textsuperscript{2},
     Wei-An Lin\textsuperscript{2},
     Hui Qu\textsuperscript{2}\\ 
     Mingi Kwon\textsuperscript{2*,3},
     Ratheesh Kalarot\textsuperscript{2}
    \\[1ex]
    \textsuperscript{1}University of Michigan
    \quad
    \textsuperscript{2}Adobe Inc. (ASML)
    \quad
    \textsuperscript{3}Yonsei University
}
\begin{document}

\maketitle


\begin{abstract}

Diffusion models have shown tremendous results in image generation. However, due to the iterative nature of the diffusion process and its reliance on classifier-free guidance, inference times are slow. In this paper, we propose a new distillation approach for guided diffusion models in which an external lightweight guide model is trained while the original text-to-image model remains frozen.
We show that our method reduces the inference computation of classifier-free guided latent-space diffusion models by almost half, and only requires 1\% trainable parameters of the base model. Furthermore, once trained, our guide model can be applied to various fine-tuned, domain-specific versions of the base diffusion model without the need for additional training: this "plug-and-play" functionality drastically improves inference computation while maintaining the visual fidelity of generated images. Empirically, we show that our approach is able to produce visually appealing results and achieve a comparable FID score to the teacher with as few as 8 to 16 steps. \href{https://5410tiffany.github.io/plug-and-play-diffusion-distillation.github.io/}{project page}
\end{abstract}

\section{Introduction}




Diffusion models \cite{ho2020denoising,sohl2015deep,song2020score} represent a novel category of generative models that have shown remarkable performance on a variety of established benchmarks in generative modeling. Specifically, conditional diffusion models \cite{rombach2022high} emerged with significantly improved sample quality by classifier-free guidance (CFG) \cite{ho2022classifier}.

However, the sampling speed of diffusion models stands out as a significant obstacle to their adoption in practical scenarios \cite{saharia2022image}. Specifically, the process of iteratively reducing noise in images typically requires a considerable number of iterations, posing challenges for efficient execution. 
For example, even when using widely adopted state-of-the-art diffusion models such as Stable Diffusion~\cite{rombach2022high}, more than 20 denoising steps are required to generate high-quality images. Moreover, when applying classifier-free guidance, two forward passes — one for the conditioned and another for the unconditioned diffusion model — are needed per denoising step, further increasing the computational cost.

\begin{figure}[t!]
    \centering
    \includegraphics[width=\linewidth]{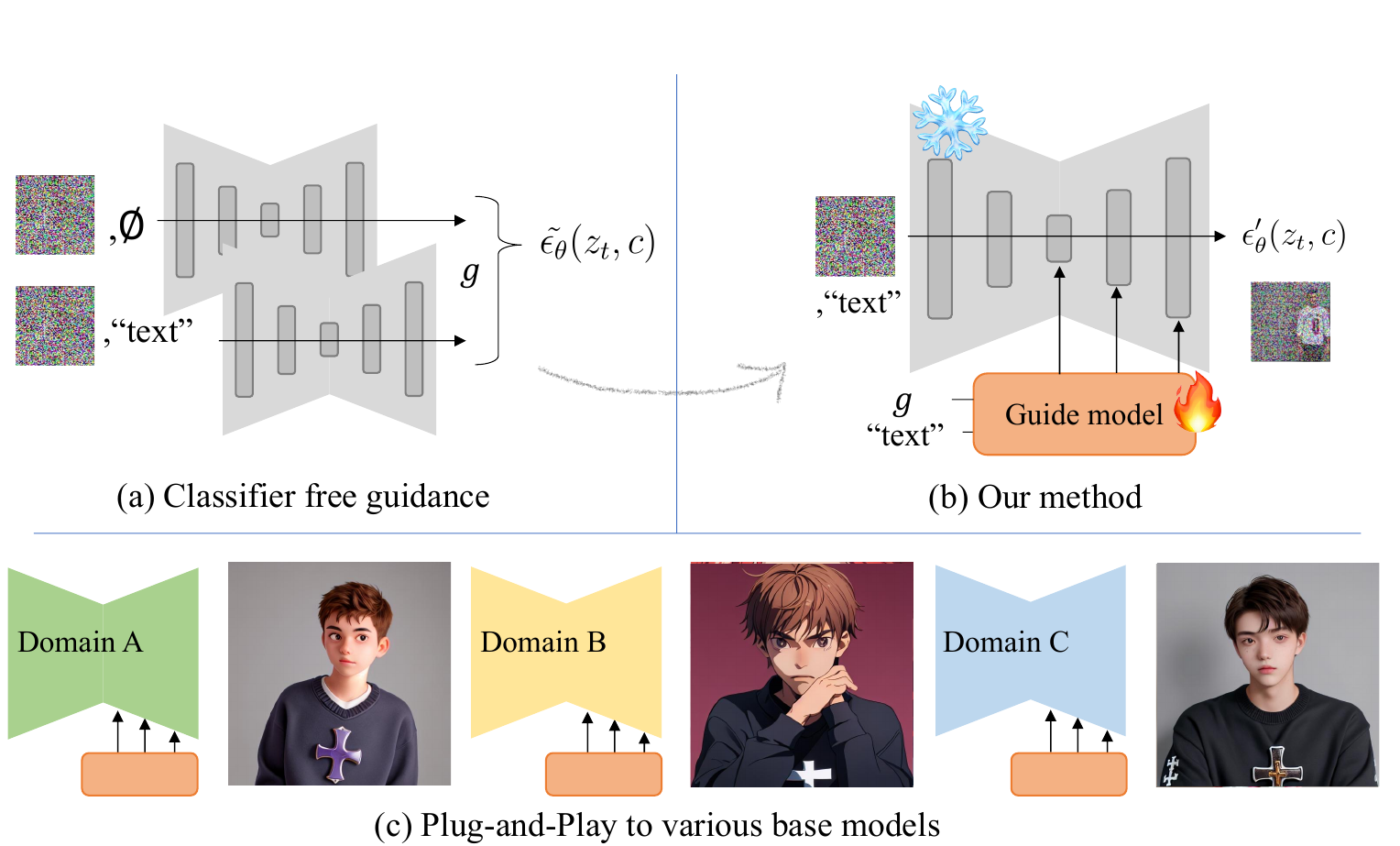}
    \caption{We trained a guide model to replace classifier-free guidance that can be plug-and-play to other base models with different domains.}
    \label{fig:concept}
\end{figure}    
\begin{figure*}[h!]
    \centering
    \includegraphics[width=0.9\linewidth]{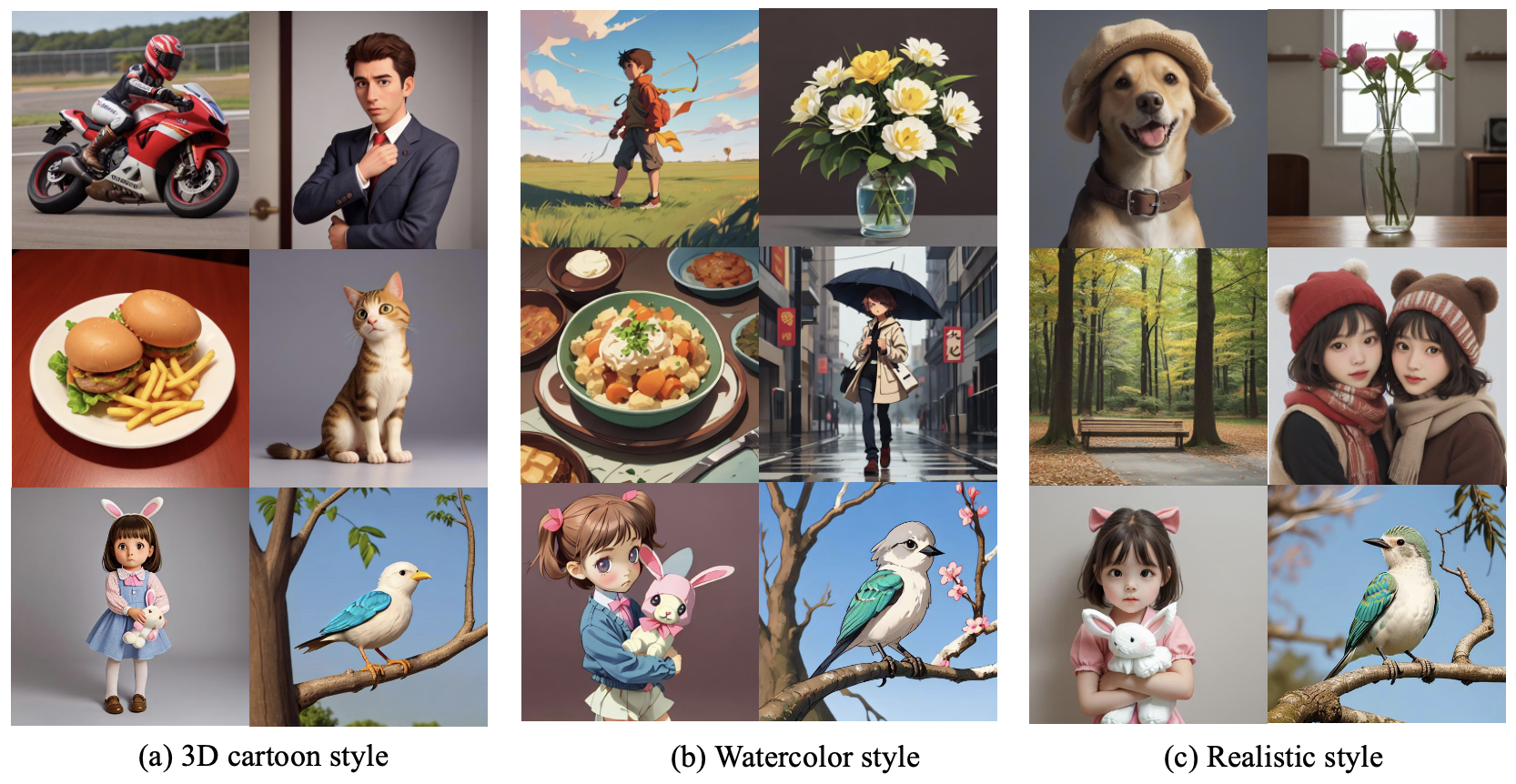}
    \caption{\YT{Applying our trained guide model to different fine-tuned latent diffusion models (LDM).}}
    \label{fig: diff-domain}
\end{figure*}

One standard approach to address the speed issues is through distillation, where a student model, initialized with the weights of the teacher diffusion model, is trained to regress to the output of the teacher that runs for multiple denoising steps~{\cite{lu2022dpm, salimans2022progressive}. However, the standard diffusion distillation approach has the following limitations. First, the number of trainable parameters of the student model is the same (or comparable) as that of the teacher diffusion model. But recent state-of-the-art diffusion models such as Imagen \cite{saharia2022photorealistic}, eDiff-I \cite{balaji2022ediffi}, and SDXL \cite{podell2023sdxl} often have billions of parameters, and distilling these large models requires a tremendous amount of computation. Second, it has been shown that diffusion models can be finetuned to different domains. When finetuned on a collection of customized images, diffusion models can be adapted to generate content with novel structures and aesthetic styles. When finetuned with only a few images, prior work has shown that novel concepts can be learned \cite{ruiz2023dreambooth}. However, with standard distillation on the base model, these finetuned models are no longer applicable. Re-training on the distilled student model is required for all the domains of interest.

In this paper, we propose a plug-and-play distillation approach to address these issues. Specifically, we introduce a novel type of distillation that makes the parameters of the base model remain untouched: we propose an external guide model with a lightweight architecture that injects feature maps to enable the diffusion model to generate text-conditioned images on one path.


We first experiment with distilling CFG into one forward pass, which effectively reduces the inference FLOP counts by 32\%. We further study different architectural choices of the lightweight module and show that the proposed architecture is around 1\% of the parameters of the base, thus effectively halving the inference FLOP counts. Finally, we experiment with the generalizability of the plug-and-play module. Once our lightweight guided module is trained, it can be readily integrated with existing finetuned diffusion models, requiring minimal to no further training. 

In summary, our approach has the following advantages:
}

\begin{itemize}
    \item \textbf{Low computational cost for training}:
        The parameters required to learn from the distillation approach is only 1\% (42\% for the Full guide model) of the diffusion model, making the computational cost for training very low compared to other distillation methods.
   \item  \textbf{Maintaining the weights of the base model}: Our approach maintains the conditioned diffusion model as-is, maintaining the integrity of the diffusion model.
    \item \textbf{Reducing inference time}: Our approach is able to decrease the FLOPs count for each sampling step by half, and is able to produce high-quality images with only 8 steps.
    \item \textbf{Generalizability}: After the trained model is obtained, it can adapt to different types of fine-tuned base models without retraining. 
\item \textbf{Adaptable with other distillation techniques}: The model can be applied to different approaches such as progressive distillation, for further sampling steps reduction.
\end{itemize}

\section{Related work}

\begin{figure*}[th]
    \centering
    \includegraphics[width=0.9\linewidth]{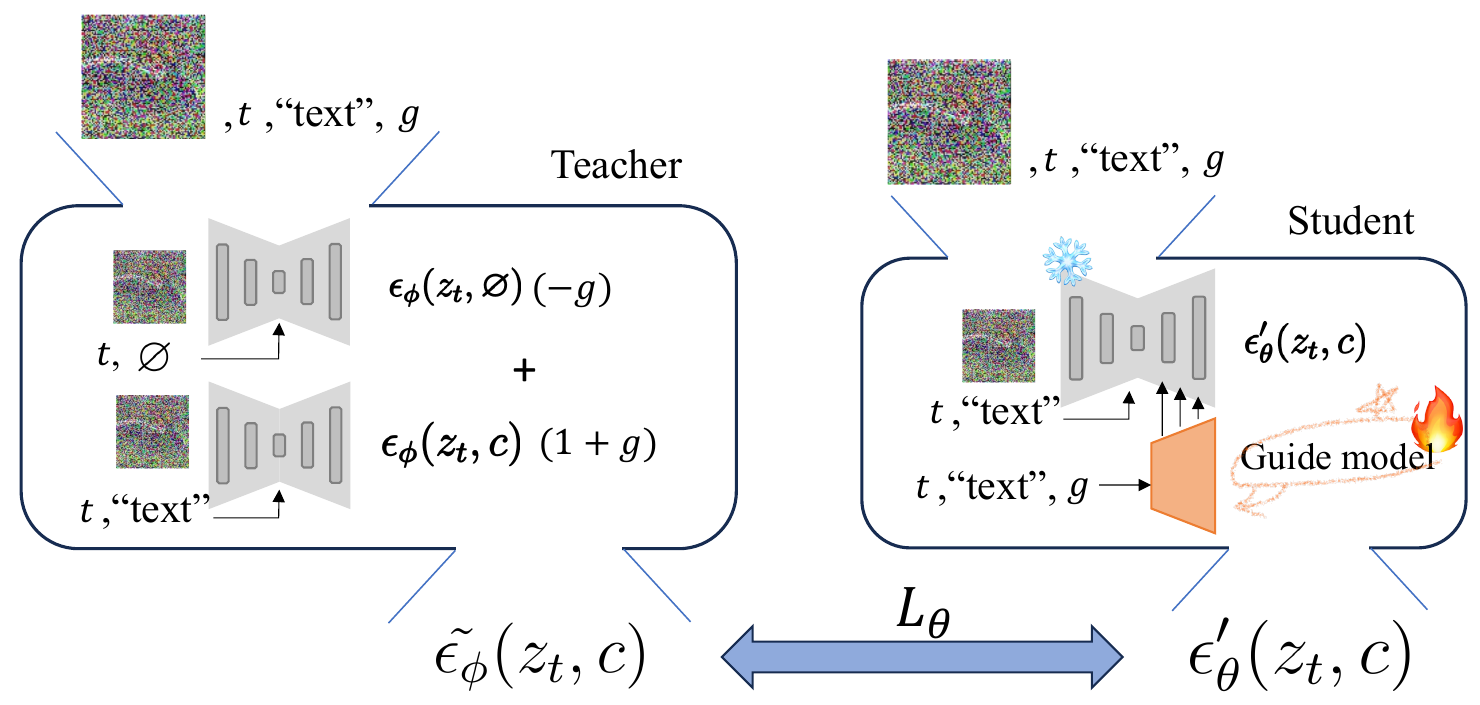}
    \caption{The overview of CFG distillation. Instead of using \YT{two} feed-forward pass and classifier-free guidance, we train a student model conditioned with the guidance, cross-attention, and time embedding to predict the output image with only one forward pass.}
    \label{fig:overview}
\end{figure*}

\subsection{Reducing inference time in diffusion models}

{To reduce the expensive computational cost in inference time of diffusion models, previous papers have attempted to improve the sampling speed of diffusion models.}

{One of the straightforward ways is designing accurate ODE samplers~\cite{lu2022dpm, liu2022pseudo, jolicoeur2021gotta}. For example, Denoising Diffusion Implicit Models (DDIM) \cite{ho2020denoising} uses first-order Euler's method which enables to reduce the inference timesteps.}



{On the other hand, there are previous works that attempt to incorporate distillation techniques to improve model inference efficiency.}
Distillation in deep learning refers to a process where a larger and more complex model (referred to as the ``teacher" model) is employed to train a smaller and simpler model (the ``student" model). The goal of distillation is to transfer the knowledge and information captured by the teacher model to the student model, enabling the student model to achieve similar performance with reduced complexity and computational requirements. \citet{golnari2023selective} proposed optimizing specific denoising steps by restricting noise computation to conditional noise and eliminating unconditional noise computation, thus reducing the complexity of target iterations. \citet{salimans2022progressive} and \citet{meng2023distillation} distilled the model to achieve fewer sampling steps. However, these methods only focus on reducing inference timesteps and require progressive model distillation, which may require more time and computing resources to train these models.

Recently, LCM-LoRA\cite{luo2023lcmlora} proposed a plug-and-play distillation approach utilizing LoRA, which has garnered significant attention. However, their method has the drawback of requiring double computation due to the use of classifier-free guidance. Our work, completed contemporaneously and accepted by CVPR, addresses similar challenges. We acknowledge the impact and relevance of their contribution.

CoDi\cite{mei2024codi}, a concurrent work presented at CVPR, excels in producing high-quality images with very few steps (e.g., 1-4) across multiple tasks, including super-resolution, text-guided image editing, and depth-to-image generation. We acknowledge their valuable contribution.

\subsection{Controlling diffusion models}
{Many researchers in the field of diffusion models have demonstrated the ability to control models using methods beyond just text input. Notably, the use of external models to inject features into diffusion models has yielded impressive results~\cite{kwon2022diffusion,zhang2023adding, li2023gligen,ye2023ip}. For instance, ControlNet~\cite{zhang2023adding} proposed an external model that uses images, skeletons, edge maps, etc., as conditions to generate corresponding images. GLIGEN~\cite{li2023gligen} successfully created desired objects within specific bounding boxes. IP-adaptor~\cite{ye2023ip} introduced a method for generating images similar to a given image condition. These approaches all successfully manipulated images by injecting values into features through external models. However, these methods have focused on conditional image generation or editing, with no instances of applying them to distillation.}

\section{Preliminary}

\subsection{Background on diffusion models}
{
Under the continuous time setting, where $t\sim Uniform [0,1]$,
the goal of a denoising diffusion model is to train a model $\epsilon_\phi$ that approximates noise given the diffused noisy real data $x \sim p_{\text{data}}$:}
\begin{equation}
\quad \mathbb{E}_{t, \epsilon,x} [\omega (\lambda_t)||\epsilon_\phi(x_t)-\epsilon||^2_2]
\end{equation}
where $\omega (\lambda_t) = \omega (\log\left(\alpha_t^2/\sigma_t^2\right)$) is a pre-defined weighted function that takes into the signal-to-noise ratio $\lambda_t$, which decreases monotonically with $t$. $x_t$ is a latent variable that satisfied $x \sim q(x_t|x) = \mathcal{N}(x_t; \alpha_t x, \sigma_t^2 I)$. 

After training the model $\epsilon_\phi$, during the sampling stage, $x_t$ can be obtained by applying the SDE / ODE solver. For example, using DDIM:
\begin{equation}
x_t = \alpha_t \epsilon_\phi(x_t) + \sigma_t \frac{x_t - \alpha_t \epsilon_\phi(x_t)}{\sigma_t}, \quad s = t - \frac{1}{N}
\end{equation}
where $N$ is the total number of sampling steps and $x_1 \sim \mathcal{N}(0, I)$

\subsection{Classifier free guidance}
Classifier-free guidance \cite{ho2022classifier} proves to be a highly effective strategy for significantly enhancing the quality of samples in class-conditioned diffusion models. {It adopted an unconditioned class identifier $\varnothing$ as a substitute for a separate classifier that is traditionally required to create a Gaussian distribution tailored to a specific class.\cite{dhariwal2021diffusion}} This approach finds widespread application in extensive diffusion models, including notable examples like DALL·E2 \cite{ramesh2022hierarchical}, GLIDE \cite{nichol2021improved}, and Stable Diffusion \cite{rombach2022high}. In particular, Stable Diffusion designs the diffused forward and reverse process in the VAE latent space, $z=E(x), x=D(z)$ where $E$ and $D$ denote the VAE encoder and decoder. In the process of generating a sample, classifier-free guidance carries out evaluations on both conditional score estimates and unconditional score estimates. Specifically, the computation of the noise sample $\tilde{\epsilon_{\phi}}(z_{t}, c)$ follows the formulation
\begin{equation}
\tilde{\epsilon_{\phi}}(z_{t}, c) = (1 + g) \epsilon_{\phi}(z_{t}, c) - g \epsilon_{\phi}(z_{t},  \varnothing),
\end{equation}
where $\epsilon_{\phi}$ is the score estimate function that is a parameterized neural network (U-Net). $\epsilon_{\phi}(z_{t}, c)$ represents the text-conditioned term, while $\epsilon_{\phi,}(z_{t},  \varnothing)$ corresponds to the unconditional term (null text). The parameter $g$ stands for the guidance value that scales the perturbation. In this paper, we use the Stable Diffusion's VAE latent and omit the notation of Encoder and Decoder of VAE for brief. 



\section{Methodology}

\subsection{Overview}
{Inspired by ControlNet \cite{zhang2023adding}, we design the external guide network for CFG distillation by using the guidance number as the input condition. After the first stage of the distillation (i.e. CFG distillation) has been accomplished, we follow prior distillation techniques to reduce the sampling steps. This is accomplished by enabling the model to progressively learn how to halve the sampling steps~\cite{salimans2022progressive}}.
The specifics of the whole process will be elucidated in the following sections.

\subsection{CFG distillation}
    The overview of our CFG distillation method is illustrated in Figure \ref{fig:overview}.  We would like to learn a model $\epsilon_{\theta}^\prime$ to achieve
\begin{equation}
\epsilon_{\theta}^\prime(z_{t}, c; G(g, z_{t},c)) = (1 + g) \epsilon_{\phi}( z_{t}, c) - g \epsilon_{\phi}(z_{t},  \varnothing)
\end{equation}
    where $g$ is the guidance number, \guidemodel\ is our student guided model, $\epsilon_{\phi}(z_{t},  \varnothing)$ is the unconditioned U-Net forward pass, and $\epsilon_{\phi}(z_{t}, c)$ is the conditioned U-Net forward pass.  Precisely, \guidemodel\ takes the guidance as the input hint, along with time and text embedding and $z_{t}$, then injects its output feature maps to the decoder part of the original U-Net. The feature map injection can be viewed as the ``guidance strength" that helps the U-Net to trade-off between sample quality and diversity.   The pseudo algorithm is listed in Algorithm \ref{alg:s1}.


    Typically, distillation involves initializing an entirely new model that has the same structure as the teacher model and trying to make it learn the teacher's output and update the parameters of the entire student network. Instead, we use a small guide model on top of the teacher model, which leads to reduced computational overhead during training because the number of parameters in the guide models is relatively small compared to the whole U-Net. Also, this approach does not discard the teacher model after distillation training, but uses the trained guide model along with the teacher U-Net for faster inference without CFG. This feature makes it applicable to "plug-and-play" to different types of fine-tuned diffusion models directly without retraining the guide model \guidemodel.

\begin{algorithm}[!t]
\caption{CFG distillation}
\label{alg:s1}
\begin{algorithmic}
\REQUIRE real image $x$, text $c$
\STATE $\guidemodel_{\theta} \leftarrow \eta$. Initialize student guide model
\WHILE{not converged}
    \STATE Sample a timestep $t \sim Uniform[0, 1]$
    \STATE Sample a guidance number $g \sim Uniform[2, 9]$
    \STATE Sampling noise $\epsilon \sim \mathcal{N}(0, I)$
    \STATE $z_t = \alpha_t x + \sigma_t \epsilon$
    \STATE $e_{teacher} = (1 + g) \epsilon_{\phi}(z_{t}, c) - g \epsilon_{\phi}(z_{t},  \varnothing)$
    \STATE $e = \epsilon_{\theta}^\prime(z_{t}, c; \guidemodel_{\theta}(g, z_{t},c))$
    \STATE $L_\theta = \left\| e_{teacher} - e \right\|_2^2$
    \STATE $\theta \leftarrow \theta - \gamma \nabla_\theta L_\theta$
\ENDWHILE

\end{algorithmic}
\end{algorithm}

\subsection{Guide model architecture}
\begin{figure}[t!]
    \centering
    \includegraphics[width=1\linewidth]{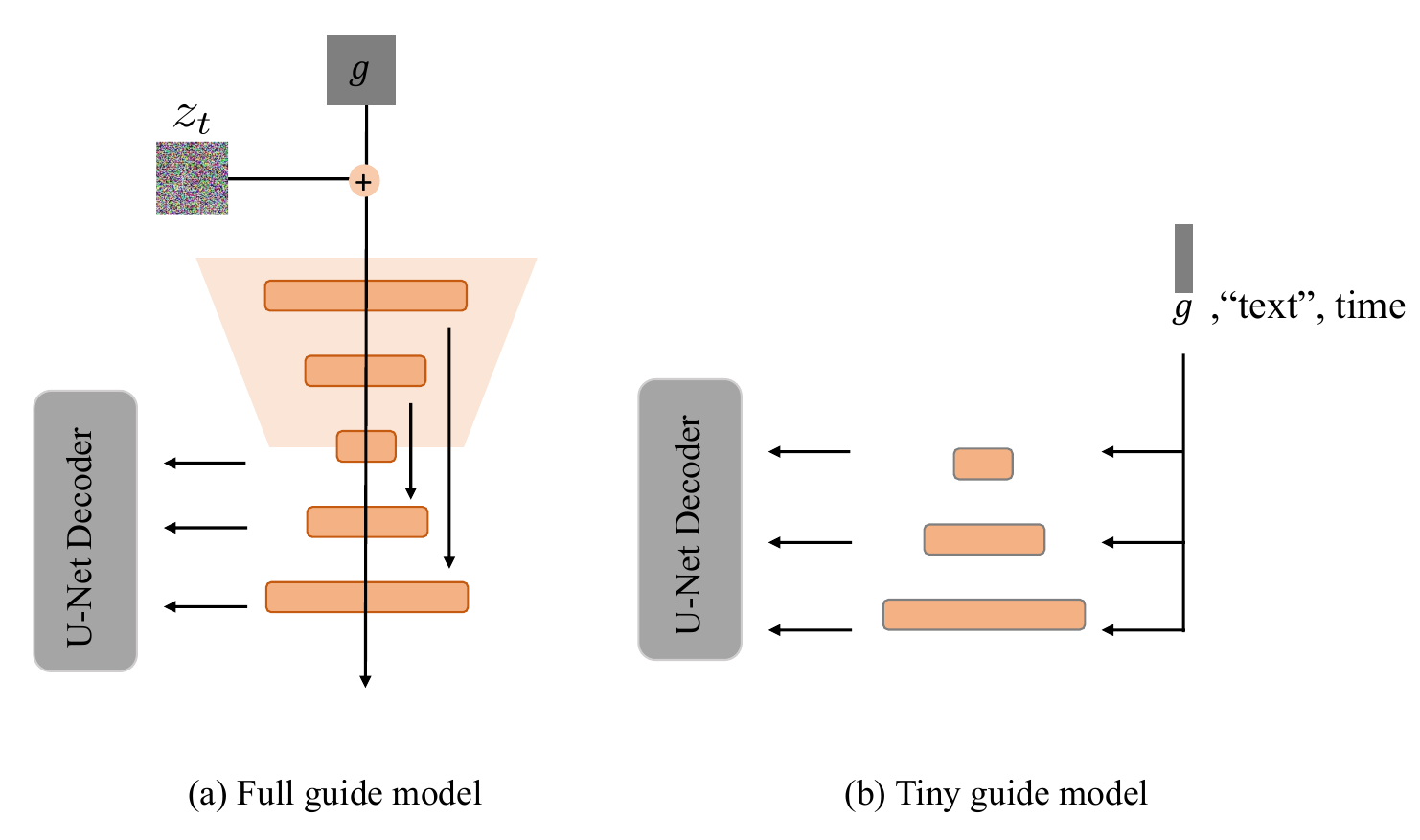}
    \caption{Comparison of the \YT{\versionone{}} architecture and the tiny architecture.}
    \label{fig:conventional vs tiny}
\end{figure}

In this section, we introduce two types of external guide model, \versionone{} and \versiontwo{}.
\paragraph{\versionone{}}
ControlNet is one of the well-designed external models for image control. When we regard the distillation with an external guide model as the external controlling, the straightforward way is using the UNet architecture of diffusion model as the guide model.
    To align with the original ControlNet architecture, our \versionone{} broadcasts the guidance number into a shape that is the same as the hint size, e.g. $(C, H, W)$. This straightforward strategy enables the model to have high capacity. The model architecture is depicted in \fref{fig:conventional vs tiny}.
    
\paragraph{\versiontwo{}}
Although \versionone{} is already a well-designed guide model, this is not an efficient way because there is not as much information needed to encode with a simple guidance number. As such, we further simplify the standard ControlNet structure, \versiontwo{}, for our guidance-distillation framework:
\YT{
\begin{equation}
y = Z\left(\gamma + Z(c_{timestep}; \Theta_{z1}) + Z(c_{text}; \Theta_{z2}); \Theta_{z3}\right)
\end{equation}
$\gamma$ is the guidance vector, which is a vector of guidance number $g$. Moreover, the timestep embedding and text embedding will also pass through zero convolution layers, denoted $Z(\cdot, \cdot)$. 
These elements are added together and passed through the zero convolutions in the decoding layer to get the corresponding output of the guide model $y$.
Zero convolution architecture ensures that undesirable noise or irrelevant features are not injected into the base model in the early stage of the training.

The \versiontwo{} simplifies the traditional ControlNet architecture by removing the encoder blocks as shown in Fig. \ref{fig:conventional vs tiny}. This design drastically reduces the number of parameters as $z_t$ no longer needs to be encoded by the guide model.

}

In following sections, we will show that our CFG distillation approach works for both the \versionone{} and the \versiontwo{}. 
\subsection{Sampling steps distillation}
After training the guide model, \guidemodel, we progressively distill it with fewer sampling steps required by incorporating with existing sampling-step-based distillation methods \cite{salimans2022progressive}.  To elaborate, under the discrete time-step scenario, let $N$ stand for the original number of sampling steps, we trained a student model to the output of two-step DDIM sampling of the teacher in one step. Precisely, the initial sampler $f(z; \eta)$ maps a random noise $\epsilon$ to samples $x$ requires $N$ steps, is distilled into a new sampler $f(z; \theta)$ that requires $N/2$ steps. $f(z; \theta)$ will become the new teacher so that we can learn another sampler that requires $N/4$ steps. This procedure will be repeated several times until the ideal sampling steps needed will be achieved. In this section, again, we only learn the parameters from the guide model $G$ and fix the base model (U-Net) throughout the distillation progress. The small size of the guide model enables the parameters to be learned quickly.

\section{Experiments}
\begin{figure*}[h]
    \centering
   \includegraphics[width=0.7\textwidth]{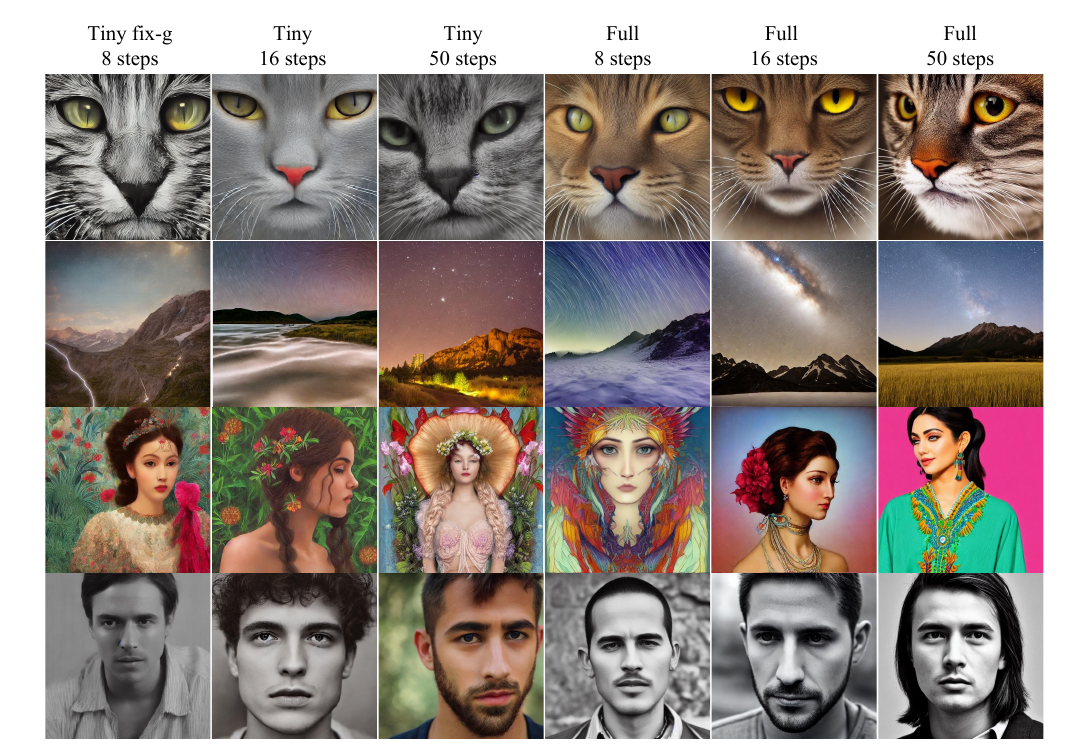}
    \caption{The results of the full guide model and tiny guide model under 8, 16, and 50 sampling steps.}
    \label{fig:results}
\end{figure*}

Distilling a diffusion model involves a balance between making the model generate images faster and maintaining good quality. Initially, we assess the image fidelity of our model through both qualitative and quantitative analyses, employing FID \cite{Seitzer2020FID} and CLIP \cite{radford2021learning} scores. Subsequently, we evaluate the effectiveness of our approach across different domains with zero additional training. Finally, since our approach keeps the original model fixed, we can closely examine latent feature maps from the guide model to better understand how guidance is applied at different timesteps during the diffusion process. 


\subsection{Setup}
We trained our model with LAION ($ 512 \times 512 $) dataset \cite{schuhmann2022laion} with the Stable Diffusion v1.5 as our score-estimation model. In the training stage, a randomly sampled guidance number $g \in [2, 9]$ is broadcast into the shape of (C, H, W), which becomes the input of the guide model. For the tiny architecture, the input is a 1d-array with the length of $C$ that passes through the zero modules along with timesteps and text embedding. We apply the $\varepsilon$-prediction model through the whole experiment. Since Stable Diffusion v1.5 is trained on $1000$ steps, we sample images with $1000$ steps as our Teacher output for our guide model to learn. We evaluate our methods with the COCO dataset \cite{lin2014microsoft}. We compare our method with DDIM \cite{zhang2022gddim} sampling and PLMS \cite{liu2022pseudo} sampling.
The FLOPs and number of parameters for our models compared to the teacher classifier-gree guidance are listed in \tref{tab:comp}. We see that our \YT{\versionone{}} only needs $\sim$ 0.67 of FLOPs of the teacher while our tiny model only computes ~ 0.51 of FLOPs of the teacher model.

\subsection{Qualitative and quantitative evaluation}
\begin{table}[!t]
\centering
\begin{tabular}{@{}lll@{}}
\toprule
\textbf{Method}        & \textbf{FLOPs (trillion)}                                       & \textbf{\# of Params (million)}                          \\ \midrule
Ours-full          & T (338.7) + 116.5                                  & T (859) + 361 \\
Ours-tiny              & T (338.7) + 7.79             & T (859) + 8.27     \\
Teacher $\times$ 2            & 677.5                                & 859                      \\ \bottomrule
\end{tabular}
\caption{The FLOP counts of single pass and number of parameters used for different architectures. Teacher $\times$ 2 stands for the dual pass FLOP counts for classifier-free guidance. T stands for the parameters or FLOPs counts of the base model.}
\label{tab:comp}
\end{table}


\begin{figure*}[t]
    \centering
    \includegraphics[width=1\linewidth]{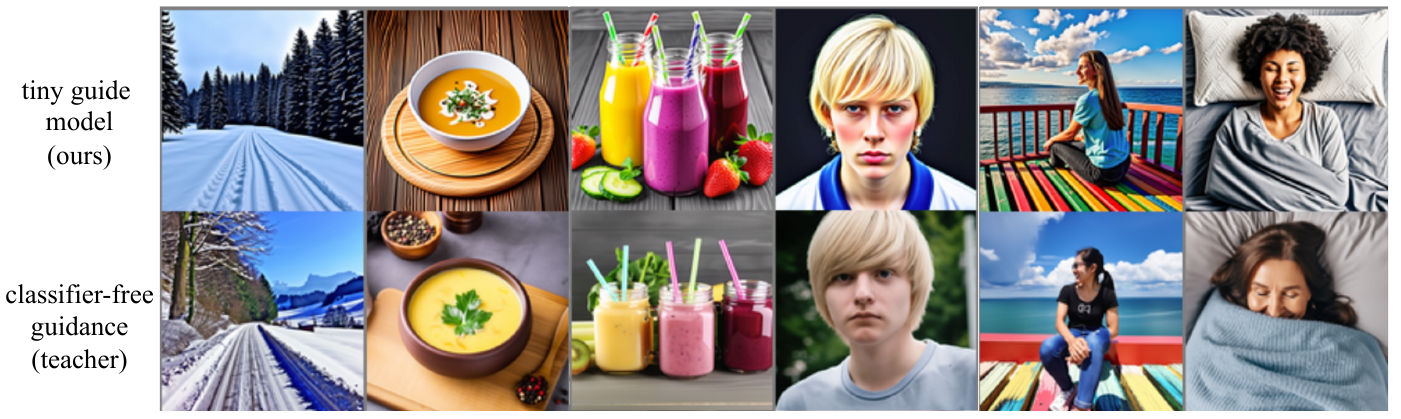}
    \caption{Disparity between the outputs of our method and classifier-free guidance with the same initial noise. Interestingly, our approach (student) usually presents a stronger contrast compared to classifier-free guidance. } 
    \label{fig: t-s-diff}
\end{figure*}

\begin{table*}[htbp]
\centering

\begin{tabular}{@{}cccccccccc@{}}
\toprule
& \multicolumn{2}{c}{\textbf{g = 2}} & \multicolumn{2}{c}{\textbf{g = 4}} & \multicolumn{2}{c}{\textbf{g = 6}} & \multicolumn{2}{c}{\textbf{g = 8}} \\ 
\cmidrule(l){2-9} 
\textbf{Methods} & \textbf{FID} & \textbf{CLIP} & \textbf{FID} & \textbf{CLIP} & \textbf{FID} & \textbf{CLIP} & \textbf{FID} & \textbf{CLIP} \\ 
\midrule
DDIM 8 $\times$ 2-step \cite{zhang2022gddim}                                    & 64.53  & 27.64 & 57.56  & 28.39 & 82.56  & 26.67 & 116.60 & 24.13 \\

Stable Diffusion v1.5 (PLMS 50 $\times$ 2 step) \cite{rombach2022high} &  17.5      &    25   &    16    &   26.63    &   18.7     &   26.50    &    21    & 26.60     \\

\midrule

Full 8step                            & 109.53 & 27.90 & 59.91  & 29.50 & 34.15  & 29.78 & 29.53  & 29.84 \\
Full 16 step                          & 49.39  & 29.15 & 31.20  & 29.70 & 21.84  & 29.92 & 20.74  & 29.91 \\
Full 50 step                               & 43.05  & 29.62 & 24.00  & 30.13 & 18.47  & 30.13 & 18.19  & 30.05     \\
Tiny 8 step                                     & 119.18 & 27.58 & 75.26  & 29.06 & 50.11  & 29.82 & 36.22  & 3023  \\
Tiny 16 step                                   & 70.66  & 28.44 & 42.06  & 29.73 & 29.19  & 30.29 & 23.14  & 30.53 \\
Tiny 50 step                                   & 52.27  & 29.19 & 32.27  & 29.81 & 28.90  & 30.08 & 19.74  & 30.23     \\
Fix guidance tiny 8 step                                    & -      &       & -      &       & -      &       & 23.97      &    30.45   \\

\bottomrule
\end{tabular}\caption{LAION dataset distillation results for text-guided latent-space diffusion models (Stable-Diffusion). We calculated the FID score by sampling 10k images from COCO 30k dataset. 
}
\label{tab:fid-scorers}
\end{table*}


\begin{figure*}[h]
\centering
    \includegraphics[width=1\linewidth]{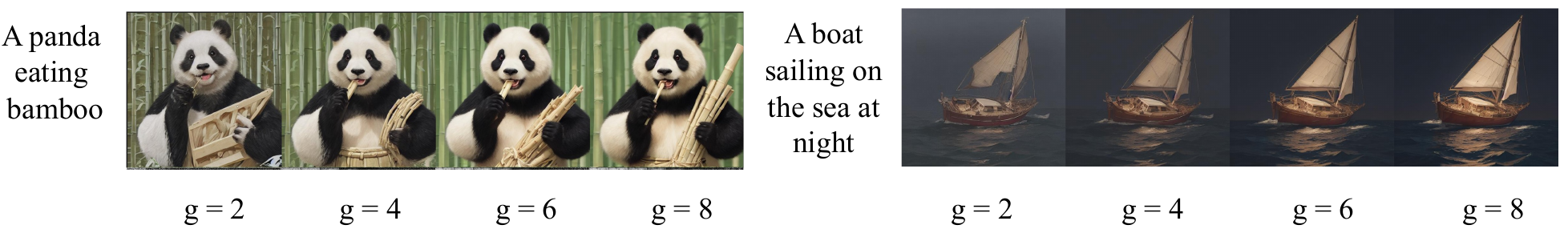}
    \centering
    \includegraphics[width=1\linewidth]{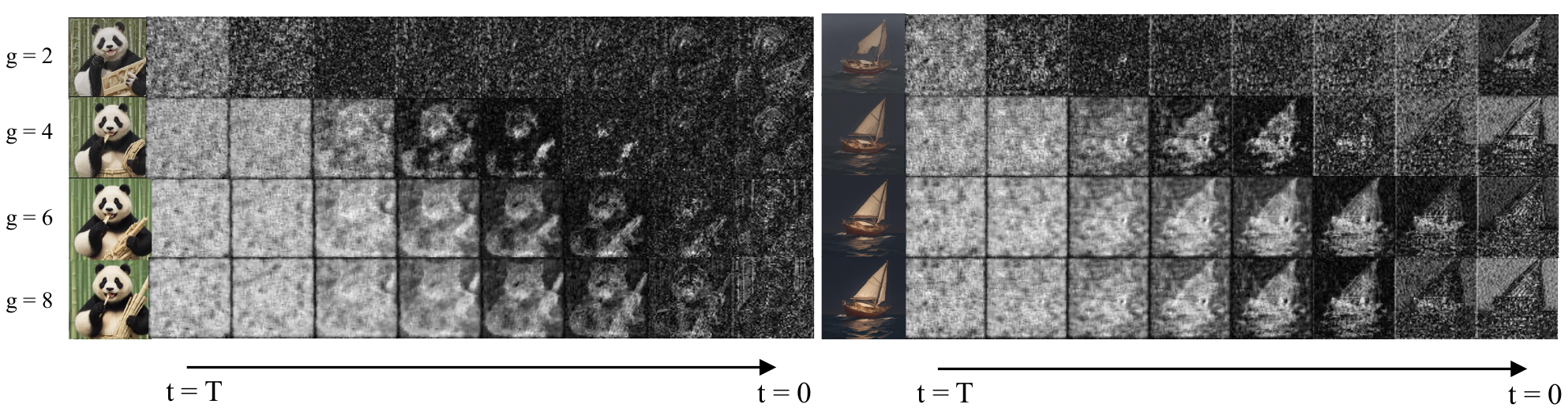}
    \caption{Visualizing the feature map injection from the guide model with guidance 8. The early stage of the iteration process has stronger injection (larger absolute values) strength, and in the later stage, the injections mainly focus on high-frequency details with lower strength. Also, the lower guidance number has lower feature map injections, higher guidance number has stronger feature map injections.}
    \label{fig:viz-diff-g}
\end{figure*}
\begin{table*}[h]
\centering

\begin{tabular}{@{}ccccccc@{}}
\toprule
                      & \multicolumn{2}{c}{\textbf{Realistic}} & \multicolumn{2}{c}{\textbf{3D Cartoon}} & \multicolumn{2}{c}{\textbf{Water Color}} \\ \cmidrule(l){2-7} 
\textbf{Score}        & \textbf{FID} & \textbf{CLIP}          & \textbf{FID} & \textbf{CLIP}           & \textbf{FID} & \textbf{CLIP}            \\ \midrule
\textbf{Ours-Tiny}      & 19.88        & 31.04                  & 22.23        & 31.07                   & 41.87        & 30.70                    \\
\textbf{CFG}      & 14.71        & 32.15                  & 18.00        & 32.11                   & 38.42        & 31.37                    \\ \bottomrule
\end{tabular}
\caption{We calculated the CLIP scores of our guide model plug into different fine-tuned models from DreamBooth without training. Guidance values are set to 8.}
\label{tab:clip-fid-generalize}
\end{table*}
\begin{figure*}[h]
    
    \centering
    \includegraphics[width= \textwidth]{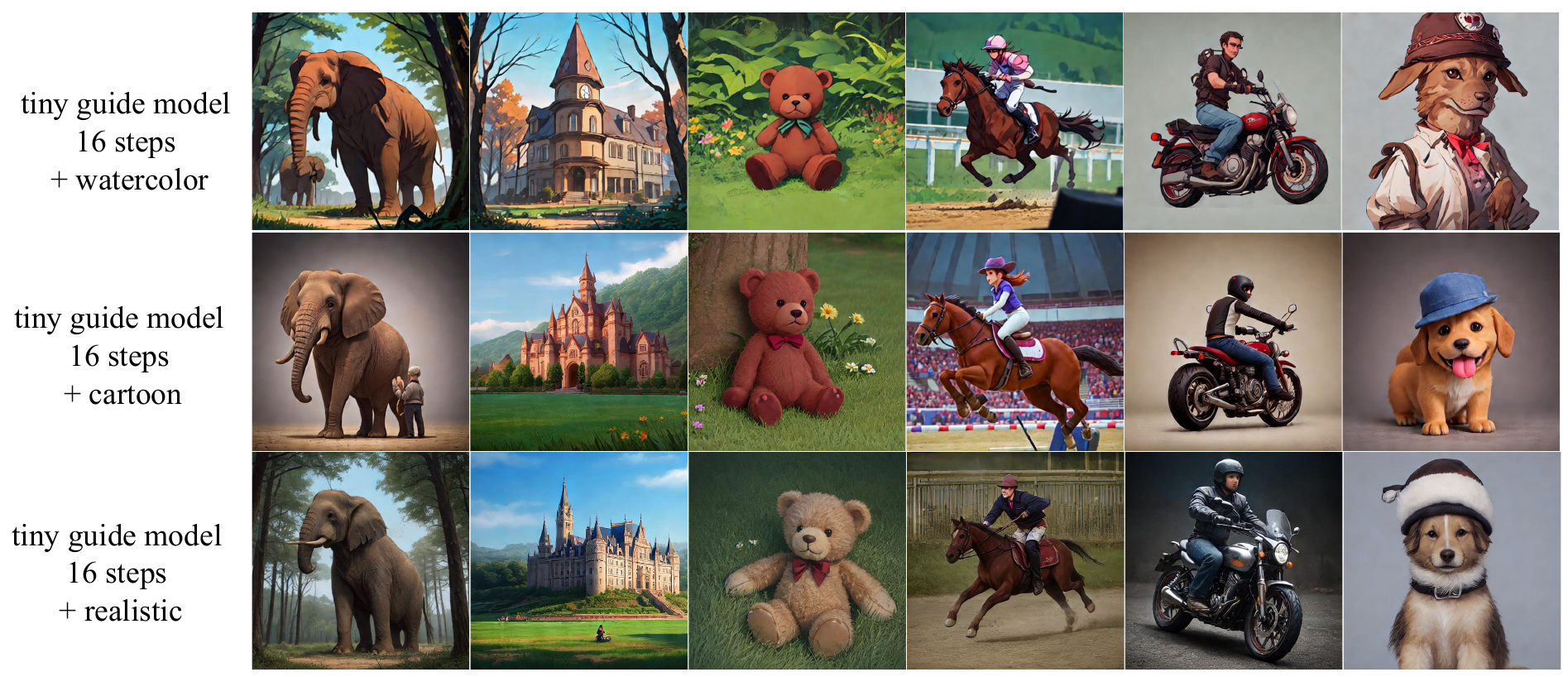}
    \caption{Plug in the tiny guide model with different fine-tuned U-Nets.}
    \label{fig:generalize}
\end{figure*}


\fref{fig: t-s-diff} illustrates the qualitative comparison between student and teacher models on various text prompts with the same initial noise. We see the quality of images generated by our guide model is close to the generated images using classifier-free-guidance while our approach can generate images and \YT{nearly} half the FLOP counts. \YT{A user study associated with images from teacher and student models can be found in the Appendix.}

\YT{Furthermore, we generate images with \YT{fewer} timesteps on \fref{fig:results}. We do not observe obvious quality degradation when decreasing our model steps to 16 and 8 \YT{with \versionone{}} given a certain level of guidance ($g$ = 8). On the other hand, since tiny guide model has less capacity, it’s challenging for the student to fully mimic the teacher’s output given a continuous guidance input during the sampling steps distillation process (i.e. progressive distillation). We observe that the \versiontwo{} can achieve almost the same image quality as \versionone{} when the sampling steps are around 50, but when the number of steps are reduced to 8, then the performance of the tiny model will degrade \YT{drastically}. This can be \YT{partially addressed} by training the tiny model with fixed guidance.  Furthermore, both of the models can achieve comparable results compared to Classifier-Free Guidance (DDIM $N \times 2$ steps). The qualitative result is shown in \tref{tab:fid-scorers}. }

\subsection{Generalizability of the guide model}
In this subsection, we show that plug in our guide model to different types of fine-tuned stable diffusion models from Dreambooth \cite{ruiz2023dreambooth} without any additional training. Our objective is to demonstrate that our guide model can acquire a general latent representation of guidance, which possesses significant adaptability to various types of fine-tuned models without training.
We focus on three different types of fine-tuned stable diffusion v1.5 models: watercolor style, realistic style, and 3D cartoon style. Then, we directly plug in our pretrained guide model \guidemodel\  to these fine-tuned models to modify their outputs. We run the models without classifier-free guidance and pass the guidance value to our guide module.
\YT{For other distillation approaches \cite{meng2023distillation}, it may be necessary to distill a new model for each domain, which can be costly in terms of training and computation}. Our approach removes this burden and make it easy to make different models finetuned for different domains \YT{nearly} two times efficiently with no additional cost.
We validate our approach by measuring FID and CLIP scores in generated images in different domains. 
\tref{tab:clip-fid-generalize} shows the FID scores and CLIP scores of the teacher CFG on these fine-tuned models versus the results of our \versiontwo{} injection approach. The generated pictures are sampled with 50 steps with guidance 8. We see that FID and CLIP scores of our tiny model which runs two times faster are comparable with CFG. In addition, qualitative results are shown in \fref{fig: diff-domain} for the \versionone{} plug-ins and \fref{fig:generalize} for \versiontwo{} plug-ins. The results indicate the great generalizability of our approach without needing to train the model for different domains.



\subsection{Latent representations of the feature map}
In this experiment, our objective is to elucidate the latent representations within the feature map injections of our guide model \guidemodel. To this end, we visualize the feature maps at various stages of the iteration process and under different guidance values. To the best of our knowledge, we are the first to visualize how classifier-free guidance emphasizes different patches of image generation in different timesteps. We are able to study this due to the architecture choice of our model that freezes the original model and adds the guide module as an additional component. By looking into feature maps of our guide module, we are able to get a better understanding of how classifier-free guidance impacts image generation.

For each layer of feature map injection, we computed the mean across various channels for each pixel and applied normalization. The number of DDIM steps used for sampling was 50.

\fref{fig:viz-diff-g} displays the feature map injections throughout the sampling process. The values indicate that the initial stages of the sampling process are more critical with respect to Classifier-Free Guidance (CFG), as this is when the primary structure of the image is formed. In the middle stage, the main subjects of the image (e.g., a panda, bamboo) are more important. Thus CFG continues to play a role in these areas, while the background becomes less significant. Finally, in the last stage of the sampling, the feature map injections mainly focus on detail refinement on the edges with low strengths (i.e. values).


Additionally, an examination of the feature maps with varying guidance values, as shown in Figure \ref{fig:viz-diff-g}, reveals a clear trend: with lower guidance, the feature map injections are less pronounced, whereas higher guidance results in more robust injections that more effectively steer the original diffusion model. Visualizations of other layers of feature maps can be found in the Appendix.



\section{Limitation}

{Although our method can significantly reduce the FLOP count in a single pass while maintaining image quality, it is important to note that, unlike CFG, our approach is not as simple to run in batch of two. It requires to run U-Net and guide module in parallel. This is a disadvantage from implementation point of view, but it is important to mention that in practice larger GPU memory consumption result in slower inference time.  
}
\section{Conclusion}

{In this paper, we introduced a method for distilling guided diffusion models \cite{ho2022classifier}. The approach allows us to efficiently train a lightweight model that modifies the outputs of the conditioned diffusion model while maintaining the base model parameters. We demonstrate that our technique substantially lowers the computational demands for latent-space diffusion models, which are classifier-free, by decreasing in the FLOP counts by half. 
Also, our method can be plug-and-play to different fine-tuned models without retraining and generate visually pleasing figures.}




{
    \small
    \bibliographystyle{ieeenat_fullname}
    \bibliography{main}
}


\clearpage
\appendix
\setcounter{page}{1}
\maketitlesupplementary

\section{More visualizations on guide model with Latent Diffusion Models}
\begin{figure*}[h]
    
    \centering
    \includegraphics[width= \textwidth]{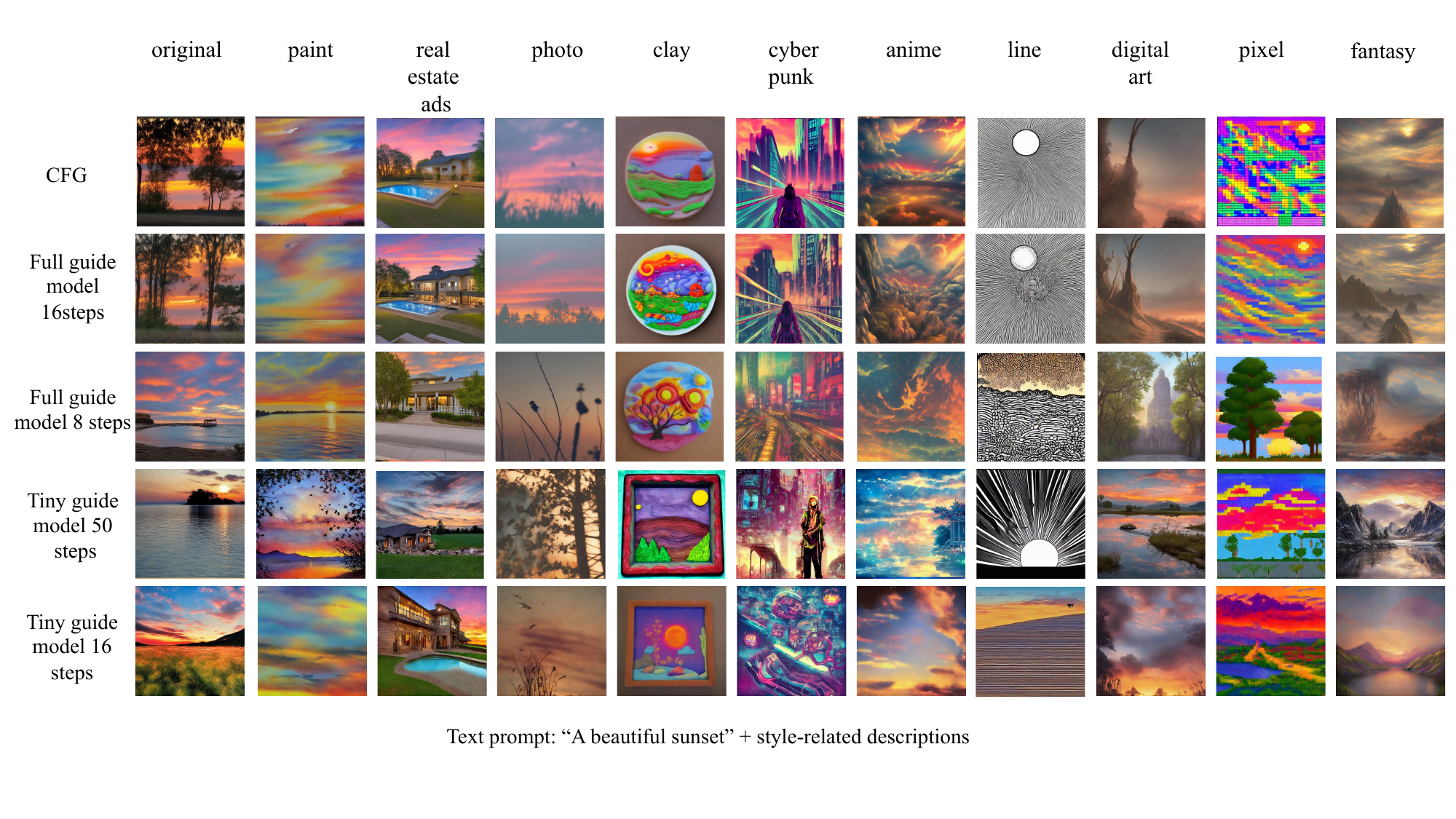}
    \caption{More visualizations on our methods working with stable diffusion v1.5. Our approach can generate different styles effectively.}
    \label{fig:more figures}
\end{figure*}

In this section, we provided more visualizations of our methods with stable diffusion v1.5. The figure is shown in \fref{fig:more figures}. This part aims to show that our approach can generate a variety of styles based on the Text prompts. Note that the initial noise in the images generated by the CFG and Full guide model with 16 steps (i.e. first two rows) is identical, but the initial noise for other methods is different.
\section{User study}
\YT{One of the characteristics observed from the injection-based conditioned model (e.g. ControlNet) is that the generated images are more saturated and have higher contrast. Some perceive them as less realistic while others may find them more visually pleasing. We conducted a user study where users were presented with a text prompt along with a pair of images generated from that text prompt (Student Full model 50 steps vs Teacher 50 steps) in a sequential fashion. They were asked to choose the preferred image based on image quality and text-image alignment. In the study, 90 participants collectively assessed a total of 680 unique text-image pairs, resulting in the accumulation of 1.8k votes. The vote distribution indicates that users did not strongly favor the teacher, with 1005 votes (55.65 \%) in favor of the teacher and 801 votes (44.35\%) in favor of the student.}

\section{Discussion on model performance with low guidance number}
\YT{
We observe that FID scores of our methods are relatively high when guidance is small (g=2, 4, 6). Due to the formulation of the guidance model, when the guidance value is small, the injection noise is small (as depicted in Figure 7). Therefore, the g=0 corresponds to not using CFG at all, which is known to generate low-quality images. 
However, when guidance is higher (g=8) our model is comparable to the teacher model.\\
}
\section{Other Layers in the Feature Maps}
In this section, we tried to display all the other feature map injection layers from our guide model. The corresponding figure is shown in \fref{fig:all-fmaps}. Generally, other layers also show that at the beginning of the sampling, the feature map injections are stronger. But there may also be some layers (6th layer, counting from top to bottom) that show an inverse trend.
\begin{figure*}[h]
    
    \centering
    \includegraphics[width= \textwidth]{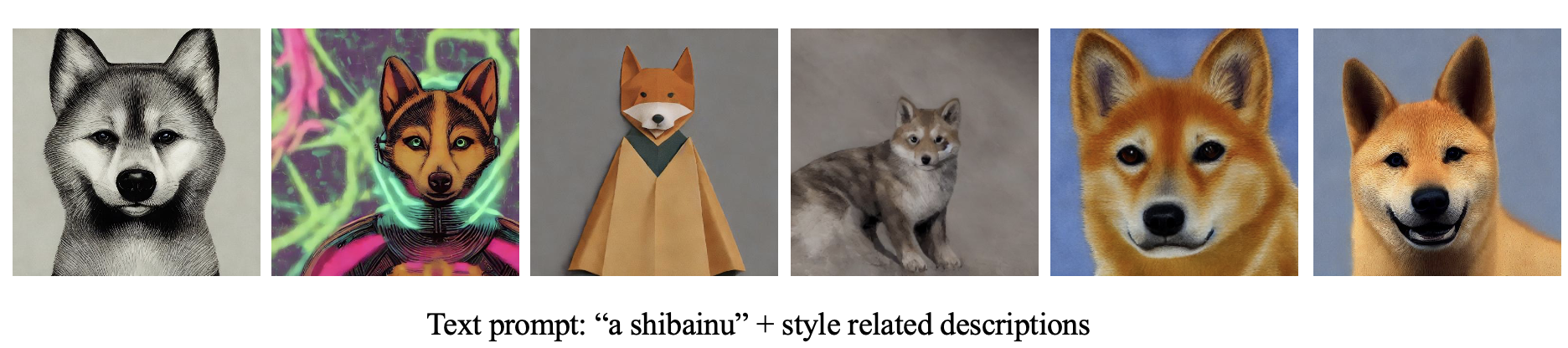}
    \caption{More results for tiny guide model 16 steps}
    \label{fig: shibainu}
\end{figure*}

\begin{figure*}[h]
    
    \centering
    \includegraphics[width= \textwidth]{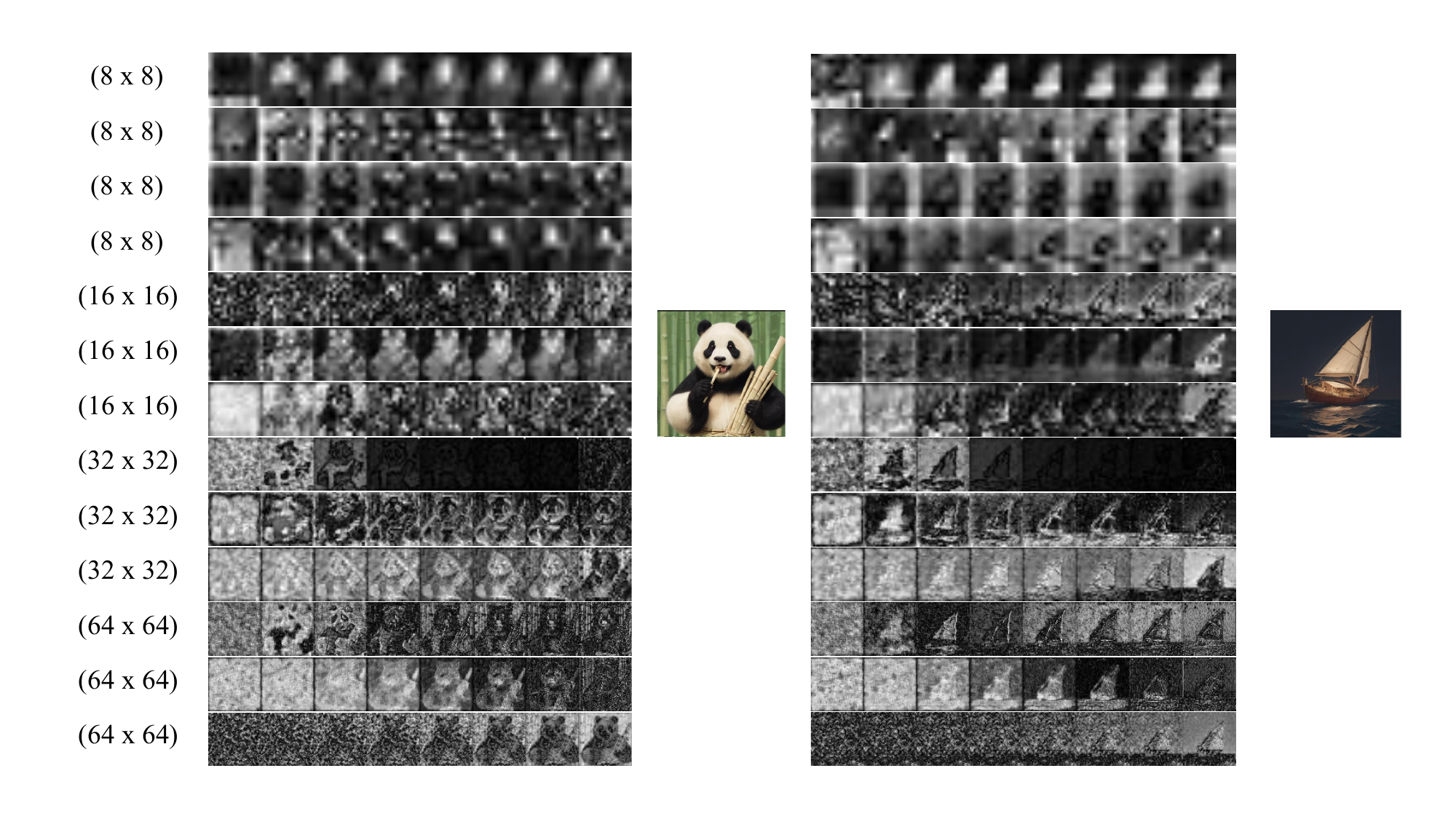}
    \caption{13 layers of feature map injections across different sampling steps. }
    \label{fig:all-fmaps}
\end{figure*}

\section{Text prompts}
In this section, we will show the precise text prompts of the generated images displayed in the main paper. The text prompt will be ordered from left to right, from top to bottom respectively.
\subsection{Text prompts in  \fref{fig: diff-domain}}
\paragraph{(a) 3D cartoon style}
\begin{enumerate}
\item A person on a racing motorcycle making a sharp right turn.
\item A businessman tying a necktie.
\item A plate of french fries and a hamburger and coleslaw.
\item A cat that is looking up while sitting down.
\item Little girl holding a stuffed bunny rabbit toy.
\item This is a bird looking in the direction of a tree.
\end{enumerate}
\paragraph{(b) Watercolor style}
\begin{enumerate}
    \item A boy walking across a field while flying a kite.
    \item An arrangement of yellow flowers with one white flower.
    \item There is a cutting board and knife with chopped apples and carrots.
    \item A woman walking under one umbrella in the rain.
    \item Little girl holding a stuffed bunny rabbit toy.
    \item This is a bird looking in the direction of a tree.
\end{enumerate}
\paragraph{(b) Realistic Style}
\begin{enumerate}
    \item A dog is wearing a fluffy hat.
    \item A vase holds green leaves and red flowers.
    \item A wooden park bench with colorful leaves on the ground.
    \item two bears giving each other a nose kiss
    \item Little girl holding a stuffed bunny rabbit toy.
    \item This is a bird looking in the direction of a tree.
\end{enumerate}
\subsection{Text prompts in \fref{fig: t-s-diff}}
\begin{enumerate}
    \item  A snow-covered road in rural environment with forest and hills in the swiss alps near schwarzenberg in the canton of lucerne, Switzerland
    \item A bowl of soup sitting on a wooden cutting board 
    \item Jars with different smoothies close-up
    \item A person with a short blond hair is looking at the camera 
    \item Happy female tourist looking sideways while sitting on colorful wooden bridge at sea viewpoint against cloud on blue sky background
    \item Satisfied forty years old European woman feels relaxed awakes early enjoys new day wears casual pajama embraces soft blanket rests long during day off 
\end{enumerate}
\subsection{Text prompts in \fref{fig:results}}
\begin{enumerate}
    \item A detailed close-up of a cat facing the camera. Its eyes are a striking feature. Vivid and expressive. The fur is meticulously rendered. Showcasing individual strands and the subtle play of light and shadow. Whiskers stand out sharply against a softly blurred background.
    \item Long-exposure night photography of a starry sky over a mountain range. with light trails. award winning photography
    \item beautiful woman wearing fantastic hand-dyed cotton clothes. embellished beaded feather decorative fringe knots. colorful pigtail. subtropical flowers and plants. symmetrical face. intricate. elegant. highly detailed. 8k. digital painting.
    \item  b\&w photography. model shot. beautiful detailed eyes. professional award winning portrait photography. Zeiss 150mm f/2.8. highly detailed glossy eyes.
\end{enumerate}
\subsection{Text prompts in  \fref{fig:generalize}}
\begin{enumerate}
    \item elephants standing on top of a grass-covered field.
    \item A castle-like building is in the background while the foreground is a green grass lawn, part of which has been mowed.
    \item A teddy bear sitting on the grass.
    \item A man performs a trick on a running horse in an enclosure
     \item The man is riding his motorcycle around the bend.
    \item A dog is wearing a fluffy hat.
\end{enumerate}






\label{sec:rationale}
 
%


\end{document}